\documentclass[10pt,twocolumn,letterpaper]{article}

\usepackage{iccv}              
\usepackage[accsupp]{axessibility}

\usepackage{times}
\usepackage{epsfig}
\usepackage{amsmath}
\usepackage{amssymb}
\usepackage[linesnumbered,ruled,vlined]{algorithm2e}
\usepackage{multibib}
\usepackage{xcolor,colortbl}
\usepackage{amsthm}

\usepackage{minitoc}
\usepackage{tocloft}

\usepackage{color}
\usepackage{booktabs}   
\usepackage{multirow}   
\usepackage{array}      
\usepackage{paralist}   
\usepackage{caption}    
\usepackage{tabularx}  
\definecolor{turquoise}{cmyk}{0.65,0,0.1,0.3}
\definecolor{purple}{rgb}{0.65,0,0.65}
\definecolor{dark_green}{rgb}{0, 0.5, 0}
\definecolor{orange}{rgb}{0.8, 0.6, 0.2}
\definecolor{red}{rgb}{0.8, 0.2, 0.2}
\definecolor{darkred}{rgb}{0.6, 0.1, 0.05}
\definecolor{blueish}{rgb}{0.3, 0.3, .6}
\definecolor{light_gray}{rgb}{0.7, 0.7, .7}
\definecolor{pink}{rgb}{1, 0, 1}
\definecolor{greyblue}{rgb}{0.25, 0.25, 1}
\definecolor{awesome}{rgb}{1.0, 0.13, 0.32}

\definecolor{figred}{rgb}{0.9, 0.1, 0.1}
\definecolor{figgreen}{rgb}{0.1, 0.7, 0.1}
\definecolor{figblue}{rgb}{0.1, 0.1, 0.9}
\definecolor{figmagenta}{rgb}{0.8, 0.1, 0.8}

\definecolor{reviewer7C2W}{RGB}{207, 49, 49} 
\definecolor{reviewerAT7d}{RGB}{34,139,34} 
\definecolor{reviewerPc7K}{RGB}{255,140,0} 

\usepackage{blindtext}

\renewcommand{\paragraph}[1]{\vspace{1em}\noindent\textbf{#1}}

\usepackage{gensymb}

\usepackage{graphicx}               

\usepackage{tabu}

\definecolor{iccvblue}{rgb}{0.21,0.49,0.74}
\usepackage[pagebackref,breaklinks,colorlinks,allcolors=iccvblue]{hyperref}
\usepackage{enumitem}

\title{MoCA: Identity-Preserving Text-to-Video Generation via Mixture of Cross Attention}

\author{Qi Xie\footnotemark[1]\hspace{0.2cm}\textsuperscript{\rm 1}~~~Yongjia Ma\footnotemark[1]\hspace{0.2cm}\textsuperscript{\rm 2}~~~Donglin Di\textsuperscript{\rm 2}~~~Xuehao Gao\textsuperscript{\rm 3}~~~Xun Yang\footnotemark[2]\hspace{0.2cm}\textsuperscript{\rm 1} \\
\textsuperscript{\rm 1} University of Science and Technology of China 
\hspace{0.3cm} \textsuperscript{\rm 2} Li Auto 
\hspace{0.3cm} \textsuperscript{\rm 3} Northwestern Polytechnical University
}

\begin{document}



\maketitle

\footnotetext[1]{Both authors contributed equally to this research.}
\footnotetext[2]{Corresponding author.}

\begin{abstract}

Achieving ID-preserving text-to-video (T2V) generation remains challenging despite recent advances in diffusion-based models. Existing approaches often fail to capture fine-grained facial dynamics or maintain temporal identity coherence. To address these limitations, we propose MoCA, a novel Video Diffusion Model built on a Diffusion Transformer (DiT) backbone, incorporating a Mixture of Cross-Attention mechanism inspired by the Mixture-of-Experts paradigm. Our framework improves inter-frame identity consistency by embedding MoCA layers into each DiT block, where Hierarchical Temporal Pooling captures identity features over varying timescales, and Temporal-Aware Cross-Attention Experts dynamically model spatiotemporal relationships. We further incorporate a Latent Video Perceptual Loss to enhance identity coherence and fine-grained details across video frames. To train this model, we collect CelebIPVid, a dataset of 10,000 high-resolution videos from 1,000 diverse individuals, promoting cross-ethnicity generalization. 
Extensive experiments on CelebIPVid show that MoCA outperforms existing T2V methods by  over 5\% across Face similarity.
\end{abstract}

\vspace{-0.25cm}
\section{Introduction}

Text-to-video (T2V) generation has made rapid progress with large-scale diffusion models~\cite{hong2022cogvideo,ho2022video,ho2022imagenvideo,chen2023videocrafter1,ma2025tuningfreelongvideogeneration, guo2023animatediff, he2022latent, blattmann2023stable, liu2024sora, lin2024open, zheng2024open,li2024hunyuan, kong2024hunyuanvideo}, enabling the creation of visually realistic and semantically coherent videos for applications in entertainment, advertising, and immersive media~\cite{xing2024survey,chen2024videocrafter2,ruan2023mm,esser2023structure,po2024state,ma2025adamsbashforthmoultonsolver}. Among these, ID-preserving T2V~\cite{he2024id,yuan2024identity,fei2025ingredients,zhang2025magic} is particularly demanding, requiring consistent facial identity across frames despite changes in pose, expression, and appearance. Therefore, achieving ID-preserving video generation with high fidelity remains a challege.

Current methods achieve identity preservation by extracting visual embeddings from a reference image and injecting them into a pretrained diffusion model \cite{rombach2022high, ho2020denoising, yang2024cogvideox, peebles2023scalable, chen2023pixart}. Despite promising progress, current controllable T2V approaches
still face key limitations: 1) \textbf{\textit{poor facial dynamics realism:}} prior face-replacement-based strategies often result in frozen or stiff expressions~\cite{he2024id, li2024personalvideo, zhong2025concat}; 2) \textbf{\textit{limited consistency between synthesis and reference:}} lacking effective facial feature extraction and integration, existing methods struggle to well maintain appearance consistency between the generated video and the reference identity image~\cite{fei2025ingredients, yuan2024identity, li2025magicid, zhang2025magic}; 3) \textbf{\textit{robustness against diverse identities:}} constrained by insufficient diversity in training data (e.g. ethnicity, gender), these methods often fail to generalize well across a wide range of identities.

To address these challenges, we propose MoCA, a Mixture of Cross-Attention Experts architecture for DiT-based Video Diffusion Models (VDMs) built upon the DiT framework. Inspired by Mixture-of-Experts (MoE) principles, our model embeds specialized MoCA layers within each DiT block to achieve inter-frame identity preservation. Each MoCA layer integrates two core components: 1) Hierarchical Temporal Pooling (HTP) that extracts identity features across multiple timescales, and 2) Temporal-Aware Cross-Attention Experts that jointly model spatial details and temporal dependencies. This dual mechanism effectively captures facial characteristics under extreme pose variations while maintaining temporal coherence. Furthermore, we introduce a Latent Video Perceptual Loss that enforces identity consistency through reference-conditional feature alignment in the latent space, preserving high-fidelity facial details across complex motion trajectories and scene dynamics. 

Unlike ID-preserving image generation, video generation lacks dedicated datasets. Existing public human video datasets are mostly limited to talking-head scenarios with constrained facial views, restricting scene and activity diversity. To address this, we introduce CelebIPVid, a high-resolution dataset of 10,000 videos featuring 1,000 diverse individuals across various ethnicities and attributes. Each video is paired with detailed captions generated by a multi-modal large language model (MLLM), and we provide five high-quality portrait images per identity to enhance reference clarity. The main contributions of our work are summarized as follows:
\vspace{-1mm}
\begin{itemize}
\item We propose MoCA, a novel mixture-of-cross-attention-experts model built on a DiT backbone, capturing facial cues across temporal scales to mitigate identity drift under large pose and motion variations.

\item We introduce Latent Video Perceptual Loss, which enforces identity consistency in latent space, preserving facial details across frames under complex motions and scene changes.

\item We construct CelebIPVid, a large-scale, diverse, and high-quality dataset for ID-preserving T2V generation research.

\item Extensive experiments on CelebIPVid show that MoCA outperforms existing T2V methods by  over 5\% across Face similarity.
    
\end{itemize}

\section{Related Work}
\label{sec:rel}

\textbf{ID-Preserved Personalization.} ID-preserving generation aims to retain specific identity attributes in synthesized images or videos. In text-to-image (T2I) tasks, diffusion models have enabled high-fidelity identity preservation by embedding identity features into generation frameworks~\cite{ye2023ip,he2024id,guo2024pulid}. However, extending this to video generation poses additional challenges, particularly in maintaining identity consistency across frames under dynamic facial motions~\cite{fei2025ingredients,li2025magicid}.
Early UNet-based methods, such as MagicMe~\cite{ma2024magic} and ID-Animator~\cite{he2024id}, leveraged fine-tuning and facial adapters to encode identity features, but often suffered from facial artifacts or identity loss in dynamic sequences. Recent DiT-based approaches like ConsisID~\cite{yuan2024identity} and Magic Mirror~\cite{zhang2025magic}, built on CogVideoX~\cite{yang2024cogvideox}, utilize cross-attention mechanisms for improved identity retention, yet still face identity drift under motion or pose changes. FantasyID~\cite{zhang2025fantasyid} introduces 3D facial geometry to preserve structure, but can lead to frame-wise semantic inconsistency or overfitting to reference features.

\textbf{Datasets for ID-Preserving Video Generation.} An ideal dataset for ID-preserving video generation should feature a large and diverse identity set, span varied scenes, focus on person-centric content, and maintain seamless temporal continuity without editing artifacts. Existing datasets include: VoxCeleb~\cite{nagrani2017voxceleb}, a talking-head dataset, includes speech and video recordings of speakers, the majority of whom are from Western countries. CelebV-HQ~\cite{zhu2022celebv} comprises 35,666 high-quality video clips ranging from 3 to 20 seconds, covering a diverse spectrum of human appearances, actions, and emotions. HDTF~\cite{zhang2021flow} contains face-centric video clips from YouTube website, with a total video duration of about 16 hours. MEAD~\cite{wang2020mead} is a multi-emotion talking-face corpus featuring 60 actors and actresses performing under 8 emotions and 3 intensity levels, designed for expression modeling and reenactment. TalkingHead-1KH~\cite{wang2021one} contains over 500,000 talking-head video clips; however, most of them are of resolution lower than 512×512, which may limit their suitability for high-fidelity generation tasks.
While person-centric, these datasets predominantly feature White and Black individuals, with limited Asian representation. Their focus on facial regions also limits broader applicability.

\section{Method}
\label{sec:meth}
\begin{figure*}[!htbp]
    \centering
    \includegraphics[width=\textwidth]{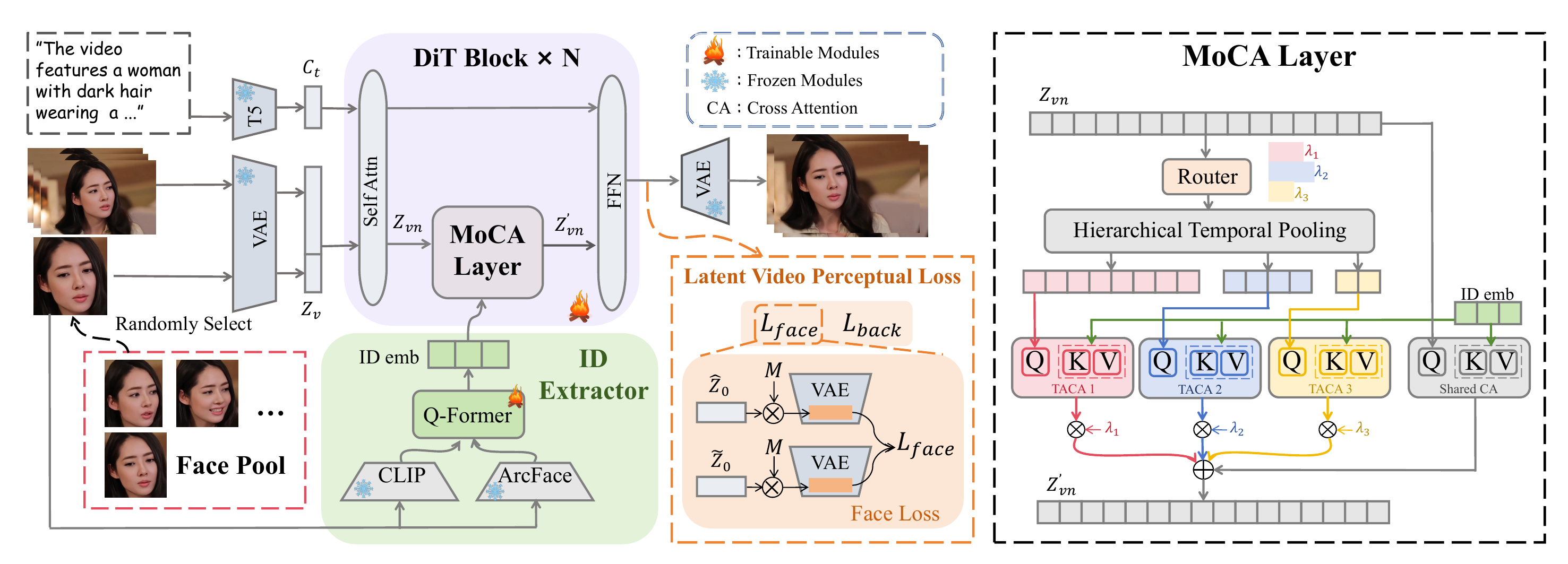}
    \vspace{-20pt}
    \caption{
    Overview of the MoCA framework for ID-preserving text-to-video generation. It incorporates Hierarchical Temporal Pooling (HTP) and Temporal-Aware Cross-Attention  (TACA) Experts to preserve identity consistency across frames. MoCA extracts global and local identity embeddings, adapting to multiple temporal scales to ensure identity coherence across varying poses and motions. Additionally, the Latent Video Perceptual Loss enforces identity consistency and high fidelity by computing the loss in latent space, focusing on both facial and background regions for overall visual quality.}
 
    \label{figure1}
    \vspace{-10pt}
\end{figure*}

\subsection{Preliminary}

Diffusion models are a powerful generative framework that learn to reverse a stochastic noise process through gradual denoising. In text-to-video synthesis, they map textual prompts to video content. Due to the high computational cost of operating in pixel space, recent methods perform diffusion in a latent spacewhere video frames are encoded via a pre-trained variational autoencoder. This improves efficiency while maintaining visual fidelity.
The optimization process can be formally expressed as follows:
\begin{equation}
    \mathcal{L}_{\text{diff\_org}} =E_{z_t,t,C_t,\epsilon}[\left\| \epsilon-\epsilon_\theta(z_t,t,C_t) \right\|^2_2]\text{,}
\end{equation}
where \( z_t \) is the latent video representation at timestep \( t \), \( C_t \) represents the textual condition at timestep \( t \), \( \epsilon \) is the noise, and \( \epsilon_\theta(z_t, t, C_t) \) is the predicted noise by the model.

\subsection{Mixture of Cross Attention (MoCA)}

\subsubsection{ID Embedding}
We continue to extract reference images from video frames, a practice validated as essential by prior works \cite{yuan2024identity,fei2025ingredients,zhang2025magic,zhang2025fantasyid}. A face detector identifies facial regions in each frame and assess whether it can be recognized by the face encoder. If the face is successfully detected, we crop the facial region and store it in a face pool. During training, one image is randomly sampled from this pool as the reference. This approach improves training robustness by leveraging diverse facial angles and expressions.
The processing involves two main components:

\noindent\textbf{Global ID Embedding Extraction} The reference image is first passed through a VAE, the same encoder used for videos, to obtain its latent representation $f_{\text{image}}$. This latent feature is then concatenated with the encoded video latent representation, forming the input $z_v$ for the DiT model.

\noindent\textbf{Local ID Embedding Extraction} Since identity information is highly complex and cannot be effectively captured by simple concatenation, we introduce an ID extractor to derive fine-grained identity features from the reference image. Specifically, the image is encoded by CLIP~\cite{radford2021learning} and ArcFace~\cite{deng2019arcface} to produce image and face embeddings respectively, which are further processed by a trainable Q-Former to yield the identity embedding $f_{\text{id}}$.

\subsubsection{MoCA Layer}
Maintaining identity consistency across frames remains a key challenge in human video generation. While base Text-to-Video (T2V) models ensure visual coherence and natural motion, they often fail to preserve fine-grained identity details, causing facial shifts during actions like head turns, eating, or speaking. Beyond modeling spatial correlations between identity embeddings and facial regions, it is also essential to capture identity-related dynamics embedded in the hierarchical temporal structure of visual tokens.
Previous ID-preserving methods mainly relied on cross-attention between identity embeddings and visual tokens, but often overlooked temporal dependencies, leading to suboptimal identity consistency. To address this, we propose a novel approach that combines cross-attention-based identity modeling with Hierarchical Temporal Pooling (HTP) and Temporal-Aware Cross-Attention (TACA) Experts to enhance identity preservation.

\noindent\textbf{Hierarchical Temporal Pooling} To help the model learn the relationship between identity embeddings and multi-level temporal dynamics, we adopt the Hierarchical Temporal Pooling (HTP) approach, applying $C$ levels of temporal pooling to process the visual tokens from DiT blocks' self-attention layers. This mechanism captures information across different temporal spans and integrates it with the identity embedding through cross-attention, enhancing identity consistency across frames. Thus, the DiT blocks contain $C+1$ cross-attention layers: one directly interacting with the original visual tokens and others processing the temporally pooled visual tokens before interacting with the identity embedding.

\noindent\textbf{Temporal Aware Cross-Attention Expert}
To incorporate extracted identity embedding into the denoising process without influencing other modalities (e.g., text embedding), we employ a cross-attention mechanism that allows visual tokens to interact with identity embedding. This design enables the model to effectively learn how identity information should be preserved throughout the generated video while ensuring that other modalities remain unaffected. This process can be mathematically expressed as:
\begin{equation}
\vspace{-3pt}
    Z_{\text{v\_org}}^\text{'} = Attention(Q, K_{\text{id}}, Q_{\text{id}})\text{,}
\end{equation}
where $Q = Z_vW_q$, $ K_{\text{id}}= f_\text{id}W_k$, $ V_{\text{id}}= f_\text{id}W_v$, and $W_q$, $W_k$ and $W_v$ are trainable parameters.

To effectively integrate identity-related information extracted at different temporal scales, we employ MoCA (Mixture of Cross-Attentions), which is inspired by MoE (Mixture of Experts). Since different DiT blocks may prioritize different temporal scales, treating all scales equally would be suboptimal. Our MoE structure consists of: (1) a shared CA, which refers to the cross-attention layer that directly interacts with unprocessed visual tokens (without HTP); (2) $C$ local temporal CAs (TACAs), which are the cross-attention layers interacting with HTP-processed visual tokens at three different scales; and (3) a router, which dynamically determines the contribution of each TACA based on the input, assigning appropriate weights. This MoCA structure ensures that the model adaptively selects the most relevant temporal features for identity preservation while keeping computational overhead manageable.

This entire process can be mathematically expressed as:
\begin{equation}
\begin{aligned}
    Z_{\text{v}}^\text{'} = Attention(Q, K_{\text{id}}Q_{\text{id}}) + \\ \sum_{i=1}^{C}{\lambda}_iAttention(H_i(Q), K^i_\text{id},V^i_\text{id})\text{,}    
\end{aligned}
\end{equation}

\vspace{-6pt}
\begin{equation}
    \mathcal({\lambda_1},{\lambda_2}... {\lambda_C}) =Router(Z_v)\text{,}
\end{equation}
where $H_i(\text{·})$ denotes the HTP operation. 

\begin{figure}[t]
    \centering
    \includegraphics[width=1.0\linewidth]{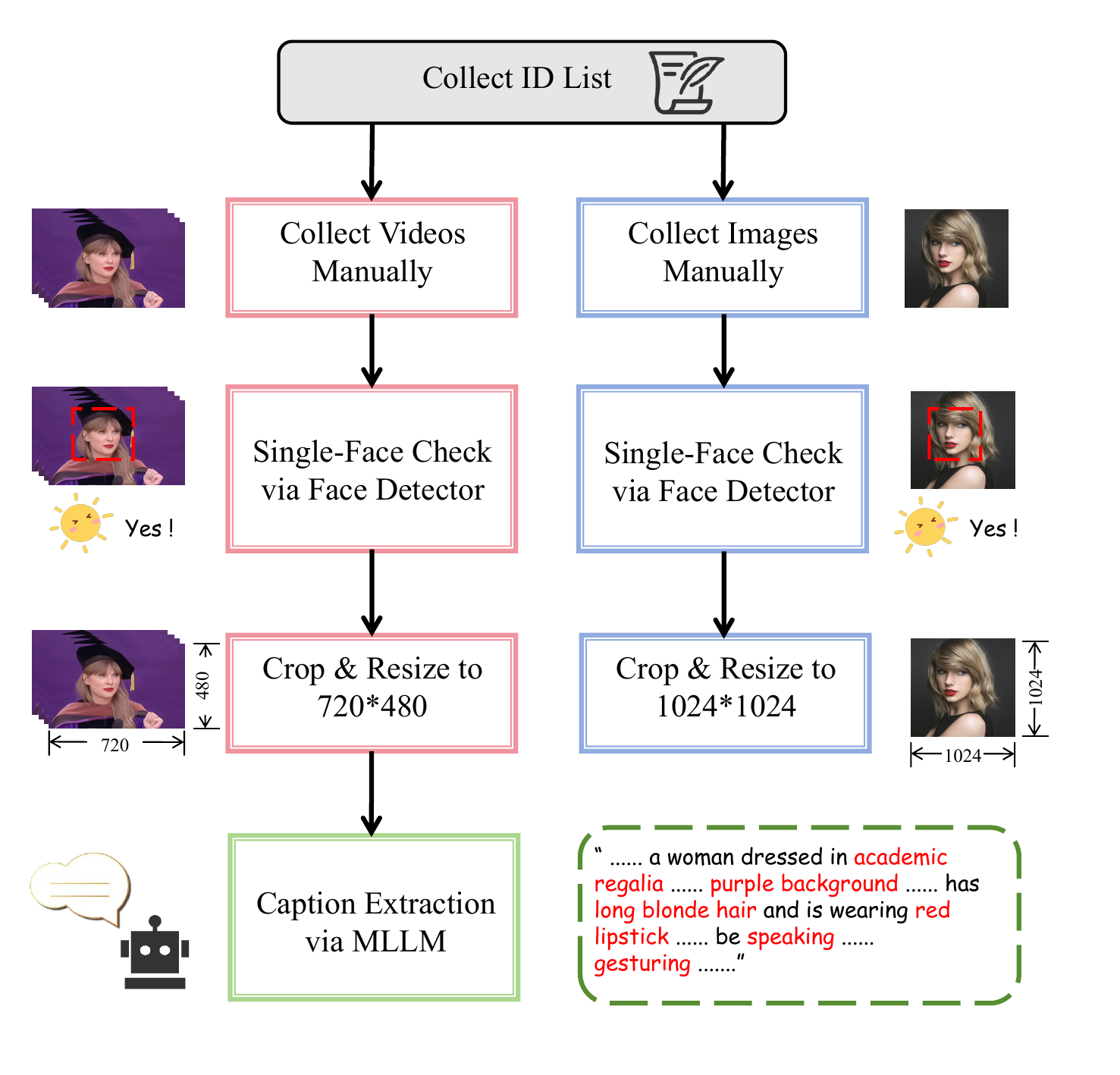}
    \vspace{-7mm}
    \caption{The data collection pipeline of the CelebIPVid.}

    \label{fig:dataset}
    \vspace{-6mm}
\end{figure}

\subsubsection{Training Strategy}

During MoCA training, the overall loss $\mathcal{L}$ combines the traditional diffusion loss with a Latent Video Perceptual Loss, enhancing identity consistency in generated videos. This encourages the model to capture key frame features while preserving temporal coherence, resulting in more realistic and consistent ID-preserving outputs.

\noindent\textbf{Diffusion Loss}
By leveraging both the global latent representation $f_{\text{image}}$ and the fine-grained identity embedding $f_{\text{id}}$ from the reference image, the objective function for diffusion is modified to:
\vspace{-3pt}
\begin{equation}
    \mathcal{L}_{\text{diff}} =E_{z_t,t,C_t,C_i,C_{\text{id}},\epsilon}[\left\| \epsilon-\epsilon_\theta(z_t,t,C_t,C_i,C_{\text{id}}) \right\|^2_2]\text{,}
\end{equation}
where $C_i$ represents the image condition, and $C_\text{id}$ represents the identity condition.

\noindent\textbf{Latent Video Perceptual Loss}
Traditional diffusion models condition the denoising of $Z_t$ on identity features from the reference image, but this affects the entire latent space---not just the face but also background, clothing, and other elements---without explicitly enforcing facial identity.
ID-Animator~\cite{he2024id} decodes $Z_0$ into video $X_0$, detects faces, and compares facial embeddings from both the generated and reference images via a face encoder, using their similarity as a reward to enforce identity. However, full decoding in DiT-based models is memory-intensive.
To mitigate this, ConsisID~\cite{yuan2024identity} computes loss directly in latent space by masking the facial region, but the use of limited facial features leads to suboptimal identity consistency.

In this paper, we propose Latent Video Perceptual Loss $\mathcal{L}p$, which enforces facial consistency in latent space without full decoding while preserving overall visual quality. It consists of two components: $\mathcal{L_{\text{face}}}$ and $\mathcal{L_{\text{back}}}$.

For $\mathcal{L_{\text{face}}}$, we extract the face mask, interpolate it into the latent space, and obtain a latent face mask. Rather than computing loss directly on masked latent features, we first pass them through the VAE decoder's initial convolutional layer, transforming them into richer feature representations before loss computation—enhancing effectiveness. The original face loss is defined as:
\begin{equation}
\vspace{-3pt}
    \mathcal{L_{\text{face}\_{\text{org}}}} = \left\| M \odot (f(\hat{Z_0}) - f(\tilde{Z_0})) \right\|\text{,}
\end{equation}
where $M$ represents latent face mask, $f(\text{·})$ represents the features extracted by passing through the initial convolutional layer of the VAE decoder.

Further more, in diffusion models, noise levels vary across timesteps. Assigning equal weight to all steps can be suboptimal, as later steps contain stronger noise, potentially dominating training and causing the model to overemphasize high-noise recovery while neglecting early-step optimization. To balance contributions across timesteps, we introduce a cosine decay factor to modulate the loss weight at each step. Specifically, the weight $w(t)$ is defined as:
\begin{equation} 
\vspace{-3pt}
w(t) = \cos \left(\frac{t}{T} \times \frac{\pi}{2} \right)\text{,}
\end{equation}
where $t$ denotes the current diffusion timestep, $T$ represents the total number of diffusion timesteps. Thus, the face loss function is updated accordingly:
\begin{equation}
\vspace{-3pt}
    \mathcal{L_{\text{face}}} = w(t)\left\| M \odot (f(\hat{Z_0}) - f(\tilde{Z_0})) \right\|\text{.}
\end{equation}

However, constraining the loss only to the face region may lead the model to overfit facial generation while overlooking other areas, compromising overall video quality. To address this, we introduce a back loss using $(1-M)$ to isolate non-face regions in both $\hat{Z_0}$ and $\tilde{Z_0}$, computing their squared difference. This encourages more holistic supervision. The initial formulation of the back loss is:
\vspace{-3pt}
\begin{equation}
    \mathcal{L}_{\text{back}\_{\text{org}}} = \left\| (1 - M) \odot (\hat{Z_0} - \tilde{Z_0}) \right\|^2\text{.}
\end{equation}

Similarly, we apply the cosine decay factor to prevent the model from overfocusing on high-noise samples:
\vspace{-3pt}
\begin{equation}
    \mathcal{L}_{\text{back}} = w(t)\left\| (1 - M) \odot (\hat{Z_0} - \tilde{Z_0}) \right\|^2\text{.}
\end{equation}
The final latent video perceptual loss $\mathcal{L}_p$ is formulated as a weighted sum of face loss and back loss.
\vspace{-3pt}
\begin{equation}
    \mathcal{L}_p = \alpha\mathcal{L_{\text{face}}} + \beta\mathcal{L}_{\text{back}}\text{,}
\end{equation}
where $\alpha$ and $\beta$ are the hyper-parameters to balance the two losses.

Finally, our overall loss function $\mathcal{L}$ is formulated as a sum of traditional diffusion loss and latent video perceptual loss, effectively improving identity consistency in generated videos.
\vspace{-3pt}
\begin{equation}
    \mathcal{L} = \mathcal{L}_{diff} + \mathcal{L}_p\text{.}
\end{equation}

\subsection{Data Collection}

Limited specialized dataset poses a significant challenge for ID-preserving video generation. In this paper, we introduce CelebIPVid, a dataset specifically designed for ID-preserving T2V training. Our CelebIPVid is built through a comprehensive data pipeline: videos are manually collected from diverse online platforms and undergo a rigorous single-face verification using a face detector before being cropped and resized to 720×480 resolution. In parallel, high-quality facial images for each identity are also manually collected—--each identity is supplemented with five images captured from different angles, with varying hairstyles and expressions, and these images are independently sourced and resized to 1024×1024 resolution. To enrich contextual information, captions are generated using a multi-modal large language model Qwen2-VL-72B~\cite{wang2024qwen2} , ensuring that each video is accompanied by a detailed textual description. The overall data collection pipeline is shown in Figure~\ref{fig:dataset}.

\begin{figure}[t]
    \centering
    \includegraphics[width=1.0\linewidth]{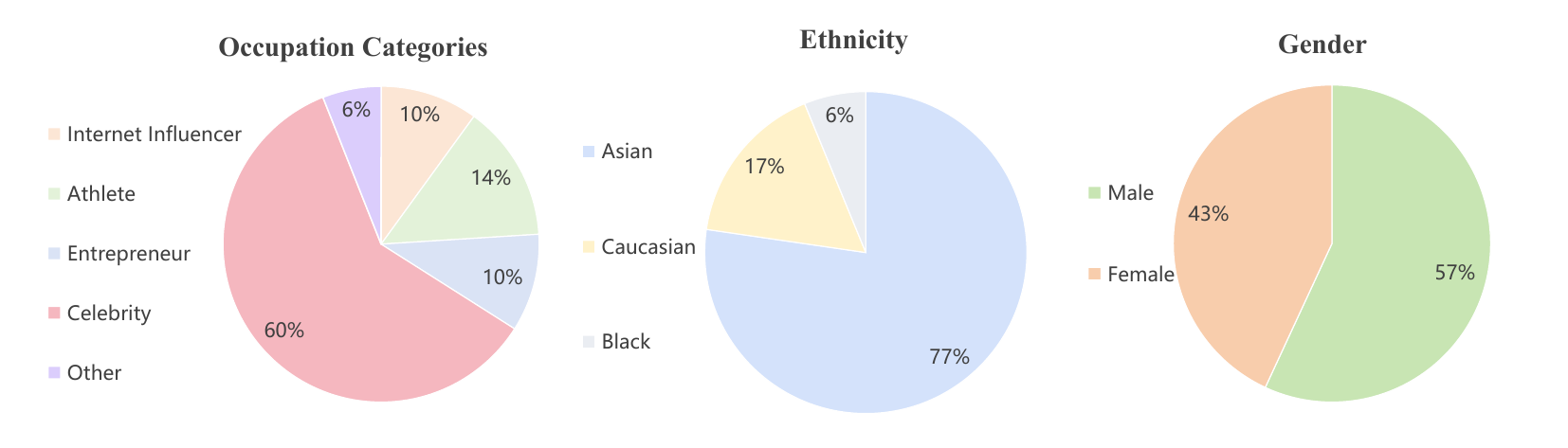}
    \vspace{-7mm}
    \caption{The data statistics of CelebIPVid.}

    \label{fig:data}
    \vspace{-4mm}
\end{figure}

Our CelebVid dataset comprises 10,000 video clips (ranging from 5 to 15 seconds per clip) representing 1,000 individuals, along with 5,000 high-quality images and corresponding text prompts. CelebIPVid is highly diverse with respect to ethnicity, gender, and professional background. For instance, as illustrated in Figure~\ref{fig:data}, the gender distribution is approximately 57\% male and 43\% female. The ethnicity distribution is 77\% Asian, 17\% Caucasian, and 6\% Black identities. Moreover, the occupation categories span a broad spectrum—including celebrities, entrepreneurs, athletes, internet influencers, and others—ensuring extensive diversity in identity representation. In this work, CelebIPVid serves as the training dataset for our MoCA model.

\section{Experiments}
\label{sec:exp}

\begin{figure*}[t]
    \centering
    \includegraphics[width=0.9\linewidth]{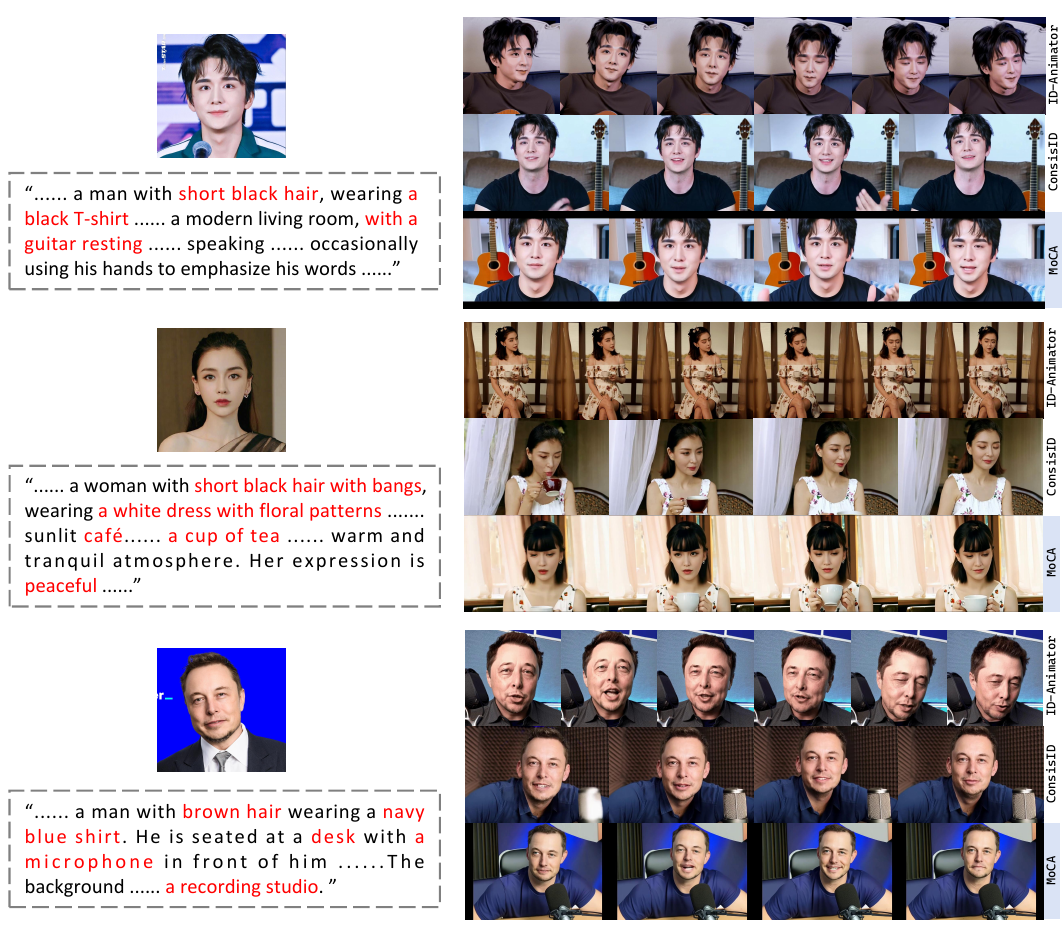}
    \vspace{-5mm}

    \caption{
 Qualitative Comparisons: Comparison of video outputs from ID-Animator, ConsisID, and our proposed MoCA across three representative cases.
    }
    
    \label{fig:compare}
    \vspace{-2mm}
\end{figure*}

\subsection{Implementation.}
We adopt CogVideoX-5B~\cite{yang2024cogvideox}, a DiT-based generative architecture, as the baseline of MoCA. For training, we utilize our self-collected dataset CelebIP-Vid, which contains 10,000 human videos covering a wide range of diverse scenes. Each video is cropped to 6 seconds (49 frames) to match the input requirements of CogVideoX~\cite{yang2024cogvideox}. MoCA is trained for 4 epochs with a batch size of 8 and a learning rate of 3e-6. Regarding hyperparameters, we set the number of experts $C$ to 3, the HTP pool sizes to [2,4,8], the face loss weight $\alpha$ to 2, and the background loss weight $\beta$ to 0.5. During inference, we use DPM sampling with 50 steps and a text-guidance scale of 6.0.

\subsection{Evaluation Metrics.}
Since there is no dedicated evaluation dataset for ID-preserving video generation, we curated a set of reference images by selecting portraits of 26 individuals (13 males and 13 females) from the internet. These images are not part of the training dataset and are used solely as reference inputs during inference. For the text prompts, we use the large language model GPT-4o to generate eight descriptive sentences for each identity, with specific instructions to include details about hairstyle, clothing, and background. 
We evaluate the generated videos using the following metrics: FaceSim-Arc~\cite{deng2019arcface} (F-Arc) and FaceSim-Cur (F-Cur) for face similarity, CLIPScore~\cite{hessel2021clipscore} (CLIP-T) for text alignment, Subject Consistency~\cite{huang2024vbench} (Sub) for identity consistency, and Imaging Quality~\cite{huang2024vbench} (Imaging) for visual quality; detailed definitions are provided in the Appendix.

\begin{table}[t]
\centering
\caption{Quantitative Comparison with baseline methods.}

\label{tab:quantitative_result}
\vspace{-2mm}
\resizebox{\columnwidth}{!}{
\begin{tabular}{l|ccccc}
\toprule
Methods & F-Arc$\uparrow$ & F-Cur$\uparrow$ & CLIP-T$\uparrow$ & Sub$\uparrow$ & Imaging$\uparrow$  \\ 
\midrule
CogVideoX-I2V \cite{yang2024cogvideox} & 0.54 & 0.53 & 24.22 & 0.903 & 0.731 \\
ID-Animator \cite{he2024id} & 0.33 & 0.35 & 28.34 & 0.973 & 0.710 \\
ConsisID \cite{yuan2024identity} & 0.55 & 0.57 & 31.23 & 0.951 & 0.729 \\
\midrule
\textbf{MoCA (Ours)} & \textbf{0.62} & \textbf{0.60} & \textbf{31.42} & \textbf{0.976} & \textbf{0.735} \\
\bottomrule
\end{tabular}
}
\end{table}

\subsection{Qualitative Analysis.}
In this section, we compare our method MoCA with ID-Animator and ConsisID, the only two open-source identity-preserving video generation methods with available code and weights. We randomly select three individuals for comparison and use ChatGPT to generate three text prompts, each with specific instructions to include descriptions of the person's hairstyle, clothing, and background. The results are presented in Figure~\ref{fig:compare}. As demonstrated, MoCA outperforms the other methods in terms of identity preservation, text alignment, and overall video quality. In Case 1, both ID-Animator and ConsisID show poor subject consistency, and the ID-Animator results fail to capture the "speaking" action mentioned in the text prompt. Case 2 shows that ID-Animator and ConsisID failed to generate the “short black hair with bangs” as instructed, indicating their over-reliance on the reference image, while only MoCA successfully achieved the desired result. In Case 3, ID-Animator causes severe facial distortion, leading to poor identity consistency, and the generated scene fails to include the "microphone" mentioned in the text prompt. ConsisID produces a simple background and does not include the "desk" as described in the prompt. In summary, MoCA outperforms the existing methods in identity preservation, text alignment, and overall video quality.
\subsection{Quantitative Analysis.}
We compare our method with existing state-of-the-art approaches. Since CogVideoX-I2V does not support the IPT2V functionality, we first use the advanced identity-preserving image generation method PhotoMaker to generate reference images based on the appearance-related descriptions in the text prompts from the test set. These images are then used as inputs to the CogVideoX-I2V model for video generation and evaluation. The quantitative results are presented in Table~\ref{tab:quantitative_result}. Our method achieves the highest scores on both face similarity metrics (FaceSim-Arc and FaceSim-Cur), demonstrating superior identity preservation. It also outperforms other methods in Subject Consistency, indicating that our approach maintains identity consistency most effectively throughout the dynamic video. Furthermore, our method achieves the highest CLIPScore, reflecting optimal alignment between the generated video and the textual description. Since the prompts in our test set contain detailed attributes such as hairstyle and hair color, the superior CLIPScore suggests that our model does not merely replicate reference images, but instead effectively incorporates and reflects textual information during video generation. Finally, our method also achieves the best performance in Imaging Quality, indicating that it produces the most visually appealing videos while preserving identity consistency, without compromising the original capabilities of the text-to-video base model.

\begin{table}[t]
\centering
    \vspace{-2mm}
\caption{Ablation results of removing MoCA layer and latent perceptual loss. }
\label{tab:ablation_result}
\vspace{-2mm}
\resizebox{\columnwidth}{!}{
\begin{tabular}{l|ccccc}
\toprule
& F-Arc$\uparrow$ & F-Cur$\uparrow$ & CLIP-T$\uparrow$ & Sub$\uparrow$ & Imaging$\uparrow$ \\ 
\midrule
w/o MoCA & 0.38 & 0.37 & 32.07 & 0.965 & 0.718 \\
w/o $\mathcal{L}_p$ & 0.42 & 0.41 & \textbf{32.10} & 0.969 & 0.727 \\
\midrule
\textbf{Full} & \textbf{0.62} & \textbf{0.60} & 31.42 & \textbf{0.976} & \textbf{0.735} \\
\bottomrule
\end{tabular}
}
\vspace{-2mm}
\end{table}

\subsection{Ablation Studies}

To evaluate the efficiency of our design components, we conduct ablation studies by considering two variants: (1) removing the MoCA layer (which integrates both Hierarchical Temporal Pooling and Temporal-Aware Cross-Attention Experts) and (2) removing the latent video perceptual loss $\mathcal{L}_p$. Quantitative results are shown in Table~\ref{tab:ablation_result}, with additional qualitative examples provided in the appendix.
 Without the MoCA layer, the model is only able to rely on the basic facial features directly encoded by the VAE, and thus fails to effectively integrate multi-level temporal features with the identity embedding. This limitation results in decreased identity and subject consistency. Similarly, when the latent video perceptual loss is omitted, the model lacks explicit constraints to focus on the facial region during training, which further degrades the identity consistency of the generated frames. Notably, the CLIPScore remains similar across the settings, indicating that our modifications do not interfere with the inherent text-to-video generation capability of the backbone model.

 \begin{table}[t]
\centering
\caption{ Hyperparameter analysis on the number $C$ of TACA experts in the MoCA layer.}
\label{tab:htp-N}
\vspace{-2mm}
\resizebox{\columnwidth}{!}{
\begin{tabular}{l|ccc}
\toprule
  & FaceSim-Arc$\uparrow$ & FaceSim-Cur$\uparrow$ & CLIPScore$\uparrow$ \\ 
\midrule
$C = 1$ & 0.51 & 0.49 & 31.04 \\
$C = 2$ & 0.54 & 0.53 & 30.71 \\
\rowcolor{blue!10}
\textbf{\boldmath$C = 3$} & \textbf{0.62} & \textbf{0.60} & \textbf{31.42} \\
$C = 4$ & 0.50 & 0.49 & 30.99 \\
\bottomrule
\end{tabular}
}
\vspace{-1mm}
\end{table}

\begin{table}
\centering
\caption{Hyperparameter analysis on the pool sizes of HTP.}
\label{tab:htp_pool_size}
    \vspace{-2mm}
\resizebox{\columnwidth}{!}{
\begin{tabular}{l|ccc}
\toprule
  & FaceSim-Arc$\uparrow$ & FaceSim-Cur$\uparrow$ & CLIPScore$\uparrow$  \\ 
\midrule
\rowcolor{blue!10}
\textbf{\boldmath$[2,4,8]$} & \textbf{0.62} & \textbf{0.60} & \textbf{31.42} \\
$[2,4,16]$ & 0.46 & 0.46 & 30.73 \\
$[2,8,16]$ & 0.48 & 0.47 & 31.11 \\
$[4,8,16]$ & 0.42 & 0.41 & 31.37 \\
\bottomrule
\end{tabular}
}
    \vspace{-2mm}
\end{table}

\subsection{Effect of Parameter Settings in MoCA Layer}
\subsubsection{Effect of the number of TACA experts}
To evaluate the impact of the number of TACA experts in MoCA layer on model performance, we conducted a hyperparameter study on the variable C, which determines how many levels of temporal features are extracted for computation. The results are illustrated in Table~\ref{tab:htp-N}, and additional qualitative examples are provided in the appendix. As shown in Table~\ref{tab:htp-N}, identity consistency reaches its lowest performance when $C=4$. This indicates that incorporating more levels of temporal features is not always better. It suggests that only certain levels of temporal information are truly beneficial for maintaining identity consistency in dynamic facial scenarios, while an excessive number of temporal scales may unnecessarily increase the model’s complexity and training difficulty. Conversely, configurations with $C=1$ and $C=2$ fail to capture sufficiently diverse temporal dynamics, limiting the ability of the Mixture of Cross-Attention architecture to integrate information across multiple temporal scales. Among all tested values, $C=3$ achieves the best trade-off between representation granularity and modeling capacity, resulting in the most optimal identity consistency.

\subsubsection{Effect of the pool sizes of HTP}
To evaluate the impact of pool size settings in Hierarchical Temporal Pooling (HTP), we conducted a hyperparameter study with four configurations: [2,4,8], [2,4,16], [2,8,16] and [4,8,16]. Specifically, a pool size of 2 extracts short-term temporal features from video latents, 4 extracts mid-term features, 8 extracts long-term dependencies, and 16 extracts global temporal information. The results are presented in Table~\ref{tab:htp_pool_size}, and additional qualitative results are provided in the appendix. Among all configurations, [2,4,8] achieves the best performance. This is primarily because the pool size of 16, which directly targets global temporal information, tends to overlook fine-grained temporal cues that are critical for preserving identity-specific details. As a result, it offers limited guidance for maintaining identity consistency. These results further suggest that incorporating lower-level temporal features is more effective for maintaining identity consistency under complex facial movements, as they are more likely to capture facial actions, which typically occur over shorter durations.

\section{Conclusion}

In this paper, we propose MoCA, a Diffusion Transformer (DiT)-based framework for identity-preserving text-to-video generation. By introducing the Mixture of Cross-Attention Experts (MoCA) mechanism and incorporating Hierarchical Temporal Pooling (HTP), MoCA effectively captures identity features across multiple temporal scales, mitigating identity drift during dynamic pose changes. Additionally, the proposed latent video perceptual loss enforces a balanced supervision strategy, preserving facial fidelity while maintaining global video quality through cosine-decayed weighting. To further support identity-preserving video synthesis, we introduce CelebIPVid, a high-resolution dataset consisting of 10,000 videos from 1,000 diverse individuals. Experiments on CelebIPVid demonstrate MoCA's ability to achieve superior identity consistency and video quality, outpacing existing methods in both quantitative and qualitative evaluations.

{
    \small
    \bibliographystyle{ieeenat_fullname}
    \bibliography{main}
}
\appendix

\twocolumn[
\centering
\Large
\textbf{Appendix of MoCA: Identity-Preserving Text-to-Video Generation via Mixture of Cross Attention} \\
(Supplementary Material)
\vspace{+1em}
] 
\appendix


\begin{enumerate}[label=\Alph*.]
  \item \textbf{CelebIPVid Dataset Samples} \dotfill \pageref{sec:celebipvid}
  \item \textbf{Details of Evaluation Metrics} \dotfill \pageref{sec:metrics}
  \item \textbf{Additional Experimental Results} 
  \begin{enumerate}[label*=\arabic*.]
    \item User Study \dotfill \pageref{sec:userstudy}
    \item Ablation Study on Latent Video Perceptual Loss \dotfill \pageref{sec:abl_lvp}
    \item Quantitative Study on Training Steps \dotfill \pageref{sec:steps}
    \item Comparison of Computational Cost \dotfill \pageref{sec:cost}
    \item Additional Qualitative Results \dotfill \pageref{sec:qual_re}
    \item[]
      \begin{enumerate}[label*=\arabic*.]
        \item Qualitative Results of MoCA \dotfill \pageref{sec:qual_moca}
        \item Qualitative Results of Ablation Study \dotfill \pageref{sec:qual_abl}
        \item Qualitative Results of Hyperparameter Study \dotfill \pageref{sec:qual_hyper}
      \end{enumerate}
  \end{enumerate}
\end{enumerate}

\section{CelebIPVid Dataset Samples} \label{sec:celebipvid}
To provide a clearer understanding of the data used in our experiments, we present visual examples from our CelebIPVid dataset, a high-resolution human video dataset containing 10,000 videos from 1,000 diverse individuals, spanning a wide range of ethnicities, genders, and personal attributes. The videos are provided at 30 frames per second with a resolution of 720×480, while associated identity images are standardized to 1024×1024 resolution.

\begin{figure*}[t]
    \centering
    \includegraphics[width=1\linewidth]{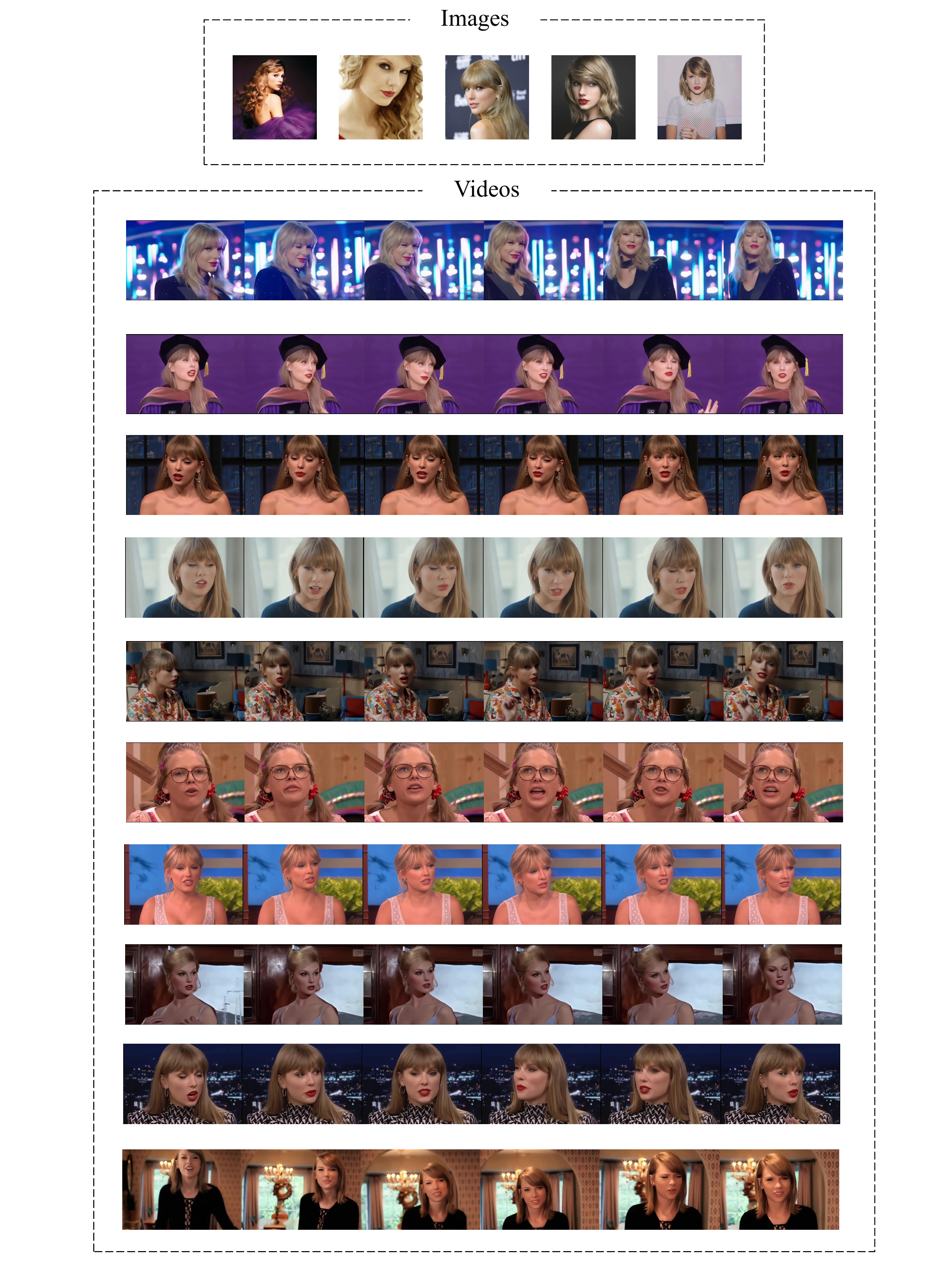}
    \caption{\textbf{A Sample from the CelebIPVid Dataset.} 
    }
    \label{fig:dataset_case}
\end{figure*}

In Figure~\ref{fig:dataset_case}, we showcase a sample from one individual in the dataset. We present the high-resolution identity image along with selected video frames to illustrate facial motion and appearance variation. This example highlights the diversity and quality of the dataset, which serves as a strong foundation for identity-preserving text-to-video (T2V) generation tasks.

\section{Details of Evaluation Metrics} \label{sec:metrics}
We use the following metrics to evaluate the quality and fidelity of the generated videos: (1) FaceSim-Arc~\cite{deng2019arcface} (F-Arc): Measures the similarity of facial features between the reference image and each frame of the generated video, as well as the consistency of facial features across frames. It uses the ArcFace embedding space. (2) FaceSim-Cur (F-Cur): Similar to F-Arc, but computed in the CurricularFace embedding space, offering an alternative facial representation for robustness.
(3) CLIPScore~\cite{hessel2021clipscore} (CLIP-T): Evaluates the alignment between the generated video and the input text prompt using CLIP-based textual similarity.
(4) Subject Consistency~\cite{huang2024vbench} (Sub): Assesses identity consistency across frames by measuring the similarity of facial appearance, clothing, and other visual attributes throughout the video.
(5) Imaging Quality~\cite{huang2024vbench} (Imaging): Quantifies the overall perceptual quality of the generated video frames based on human-like visual assessment criteria.

\section{Additional Experimental Results}

\begin{table}
\centering
\caption{ User study results for face similarity, text alignment, identity consistency, and visual quality, comparing our MoCA model with baseline methods.}
\label{tab:user_study}
\vspace{-2mm}
\resizebox{\columnwidth}{!}{
\begin{tabular}{l|cccc}
\toprule
 Methods & Face Sim$\uparrow$ & Text Align$\uparrow$ & ID Consistency$\uparrow$ & Visual Quality$\uparrow$\\ 
\midrule
CogVideoX-I2V~\cite{yang2024cogvideox} & 7.27 & 3.54 & 3.51 & 5.51 \\
ID-Animator~\cite{he2024id} & 3.11 & 4.35 & 5.21 & 4.12 \\
ConsisID~\cite{yuan2024identity} & 7.84 & 6.32 & 7.71 & 8.77 \\
\bottomrule
\end{tabular}
}
\vspace{-2mm}
\end{table}

\subsection{User Study} \label{sec:userstudy}
To better compare our method with existing state-of-the-art approaches, we conducted a human evaluation. A total of 25 participants were invited to rate 40 video clips generated by each method across four criteria: face similarity between the video frames and the reference images, text alignment, identity consistency across frames, and visual quality. Each criterion was rated on a scale from 1 to 10. The results are presented in Table~\ref{tab:user_study}. As shown, our method consistently receives higher scores across all four evaluation dimensions, further validating the superiority and effectiveness of our approach.

\subsection{Ablation Study on Latent Video Perceptual Loss} \label{sec:abl_lvp}
To evaluate the effectiveness of the Latent Video Perceptual Loss, we conducted an ablation study by removing its two components: (1) the back loss and (2) the face loss. The results are shown in Table~\ref{tab:loss}. It is worth noting that the model without the back loss (i.e., using only face loss) shows lower facial similarity compared to the full latent video perceptual loss. This is because we did not adjust the coefficient $\alpha$ of the face loss when removing the back loss. As a result, the diffusion loss takes up a larger proportion in the total loss, shifting the model’s focus toward minimizing diffusion loss and weakening the overall constraint from face loss. Consequently, facial similarity in the "w/o back loss" setting turns out to be inferior to that of the full loss.

\begin{table}
\centering
\caption{ Ablation Study on Latent Video Perceptual Loss.}
\label{tab:loss}
\vspace{-2mm}
\resizebox{\columnwidth}{!}{
\begin{tabular}{l|ccc}
\toprule
  & FaceSim-Arc$\uparrow$ & FaceSim-Cur$\uparrow$ & CLIPScore$\uparrow$ \\ 
\midrule
w/o face loss  & 0.47 & 0.45 & 31.02 \\
w/o back loss & 0.54 & 0.52 & 30.86 \\
\textbf{Full} & \textbf{0.62} & \textbf{0.60} & \textbf{31.42} \\
\bottomrule
\end{tabular}
}
\vspace{-2mm}
\end{table}

\subsection{Quantitative Study on Training Steps} \label{sec:steps}
To investigate the impact of training duration on model performance, we conduct a quantitative study across different numbers of training epochs: 2, 3, 4 (ours), and 6 epochs. As shown in Table~\ref{tab:training-step}, increasing the number of training epochs leads to significant improvements in facial similarity metrics. Both FaceSim-Arc and FaceSim-Cur rise steadily from epoch 2 to epoch 4, indicating that our model progressively learns to better preserve identity features across frames. In particular, the model trained for 4 epochs achieves the highest identity similarity scores, with FaceSim-Arc~\cite{deng2019arcface} reaching 0.62 and FaceSim-Cur reaching 0.60, substantially outperforming earlier epochs. Interestingly, while identity-related metrics improve, the CLIPScore~\cite{hessel2021clipscore} — which reflects text-video alignment — remains relatively stable across all training steps, fluctuating within a narrow range (from 30.87 to 31.91). This suggests that our model maintains robust text alignment capabilities, and does not overfit or collapse in semantic alignment even as identity fidelity increases.

In summary, training for 4 epochs achieves the best trade-off between identity preservation and semantic consistency, validating our training design choice.

\begin{table}
\centering
\caption{ Results of hyperparameter experiments on trainng steps.}
\label{tab:training-step}
\vspace{-2mm}
\resizebox{\columnwidth}{!}{
\begin{tabular}{l|ccc}
\toprule
  & FaceSim-Arc$\uparrow$ & FaceSim-Cur$\uparrow$ & CLIPScore$\uparrow$ \\ 
\midrule
epoch 2 & 0.42 & 0.41 & 31.91 \\
epoch 3 & 0.45 & 0.43 & 31.14 \\
\textbf{epoch 4} & \textbf{0.62} & \textbf{0.60} & \textbf{31.42} \\
epoch 6 & 0.55 & 0.54 & 30.87 \\
\bottomrule
\end{tabular}
}
\vspace{-2mm}
\end{table}

\subsection{Comparison of Computational Cost} \label{sec:cost}
To better evaluate the efficiency of MoCA compared to other methods, we report their computational costs during both training and inference, as shown in Table~\ref{tab:train-cost} and Table~\ref{tab:inf-cost}. As illustrated in the tables, MoCA's training time per epoch (~8 hours) is faster than other fine-tuning methods like ConsisID, demonstrating its training efficiency. During inference, MoCA is slightly faster than ConsisID. While our method introduces additional latency and memory overhead compared to the original CogVideoX-I2V baseline, we believe this is a modest and acceptable trade-off for the substantial improvement in performance. For instance, the gains in identity fidelity justify the additional computational requirements.

\begin{table}
\centering
\caption{ Training efficiency comparison.}
\label{tab:train-cost}
\vspace{-2mm}
\resizebox{\columnwidth}{!}{
\begin{tabular}{l|cc}
\toprule
  & Training Time per Epoch & GPU Memory \\ 
\midrule
ConsisID & 10h & 54GB \\
MoCA & 8h & 70GB \\

\bottomrule
\end{tabular}
}
\vspace{-2mm}
\end{table}

\begin{table}
\centering
\caption{ Inference efficiency comparison.}
\label{tab:inf-cost}
\vspace{-2mm}
\resizebox{\columnwidth}{!}{
\begin{tabular}{l|cc}
\toprule
  & Inference Time& GPU Memory \\ 
\midrule
CogVideoX & 115s & 26GB \\
ConsisID & 170s & 44GB \\
MoCA & 165s & 51GB \\

\bottomrule
\end{tabular}
}
\vspace{-2mm}
\end{table}

\subsection{Additional Qualitative Results} \label{sec:qual_re}
\subsubsection{Qualitative Results of MoCA} \label{sec:qual_moca}
\begin{figure*}[t]
    \centering
    \includegraphics[width=1\linewidth]{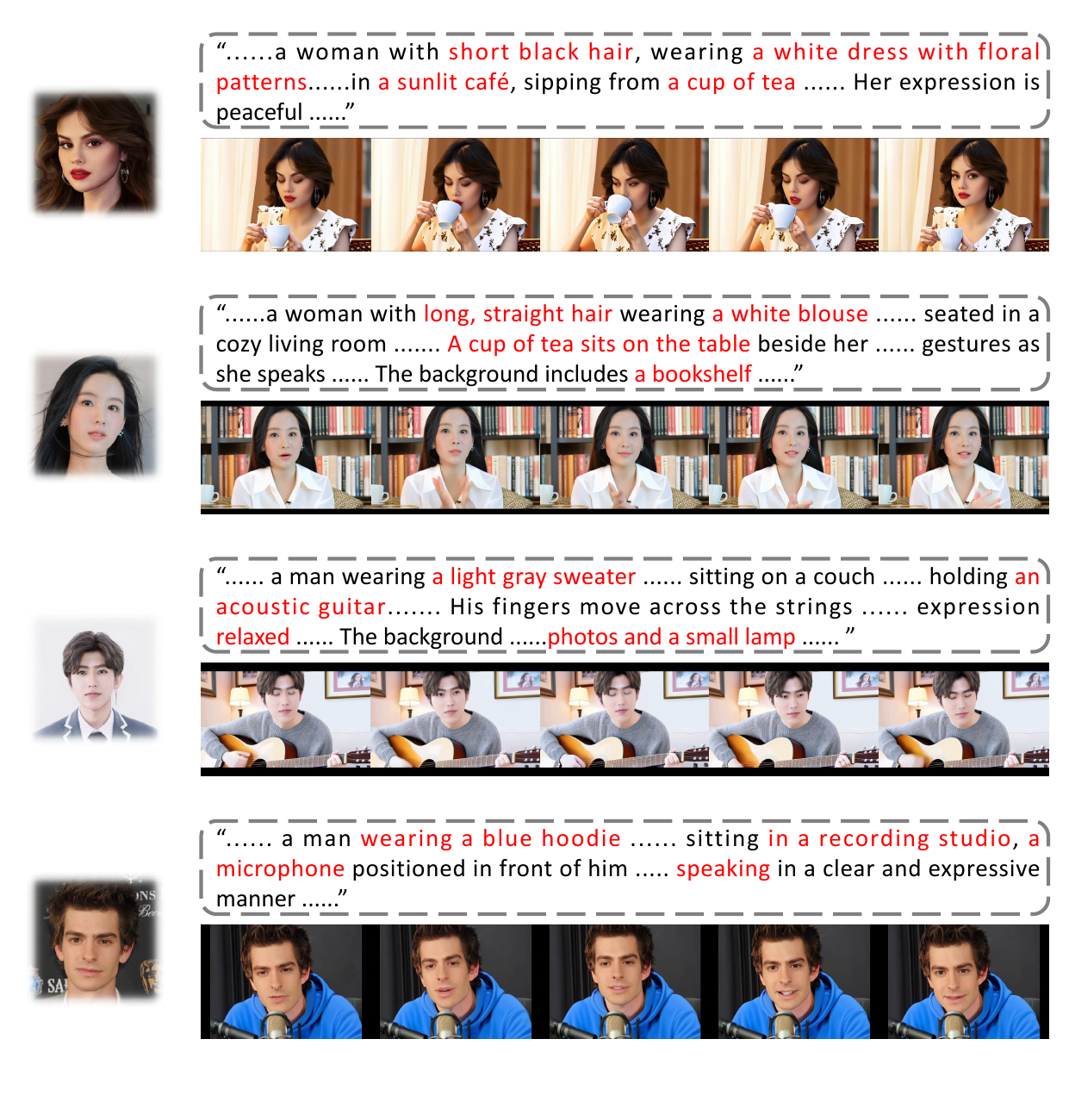}
    \caption{\textbf{Additional Qualitative Results of MoCA.} 
    }
    \label{fig:show_case}
\end{figure*}

To qualitatively evaluate the identity-preserving capability of our proposed method MoCA, we present several representative generation results in Figure~\ref{fig:show_case}. Each example consists of a high-resolution reference image that specifies the target identity, a textual prompt describing the desired motion or scene, and the corresponding video frames generated by our model.

These examples demonstrate MoCA’s ability to synthesize coherent motion while faithfully preserving the identity characteristics from the reference image. The generated results reflect not only accurate alignment with the textual instructions but also strong consistency in facial features, appearance, and style across frames.

\subsubsection{Qualitative Results of Ablation Study} \label{sec:qual_abl}

\begin{figure*}[t]
    \centering
    \includegraphics[width=1\linewidth]{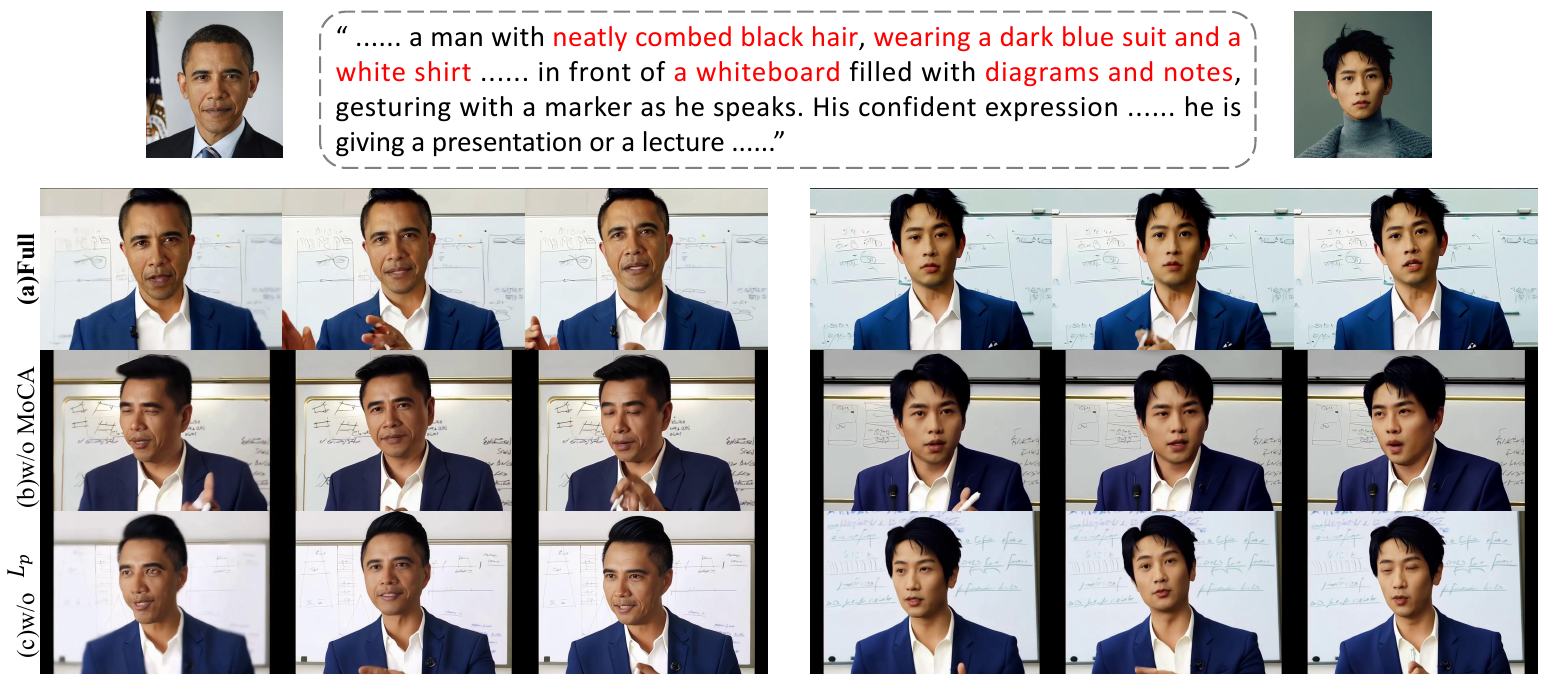}
    \caption{\textbf{Qualitative Results of Ablation Study.} 
    }
    \label{fig:wo_case_0720}
\end{figure*}

To better illustrate the effectiveness of our design components, we present two qualitative cases in Figure~\ref{fig:wo_case_0720}, comparing three conditions: (1) the full model, (2) removing the MoCA layer, and (3) removing the latent video perceptual loss $\mathcal{L}_p$. As shown in the figure, the full model produces the best results in terms of facial similarity, natural human motion, and overall video quality. When the MoCA layer is removed, the facial similarity significantly degrades, and some frames exhibit unnatural body proportions. When the latent video perceptual loss $\mathcal{L}_p$ is removed, the lack of explicit facial supervision not only leads to reduced facial similarity, but also weakens the model's ability to generate realistic faces, as evidenced by unnatural eyes in the Obama example.

\subsubsection{Qualitative Results of Hyperparameter Study} \label{sec:qual_hyper}
\begin{figure*}[t]
    \centering
    \includegraphics[width=1\linewidth]{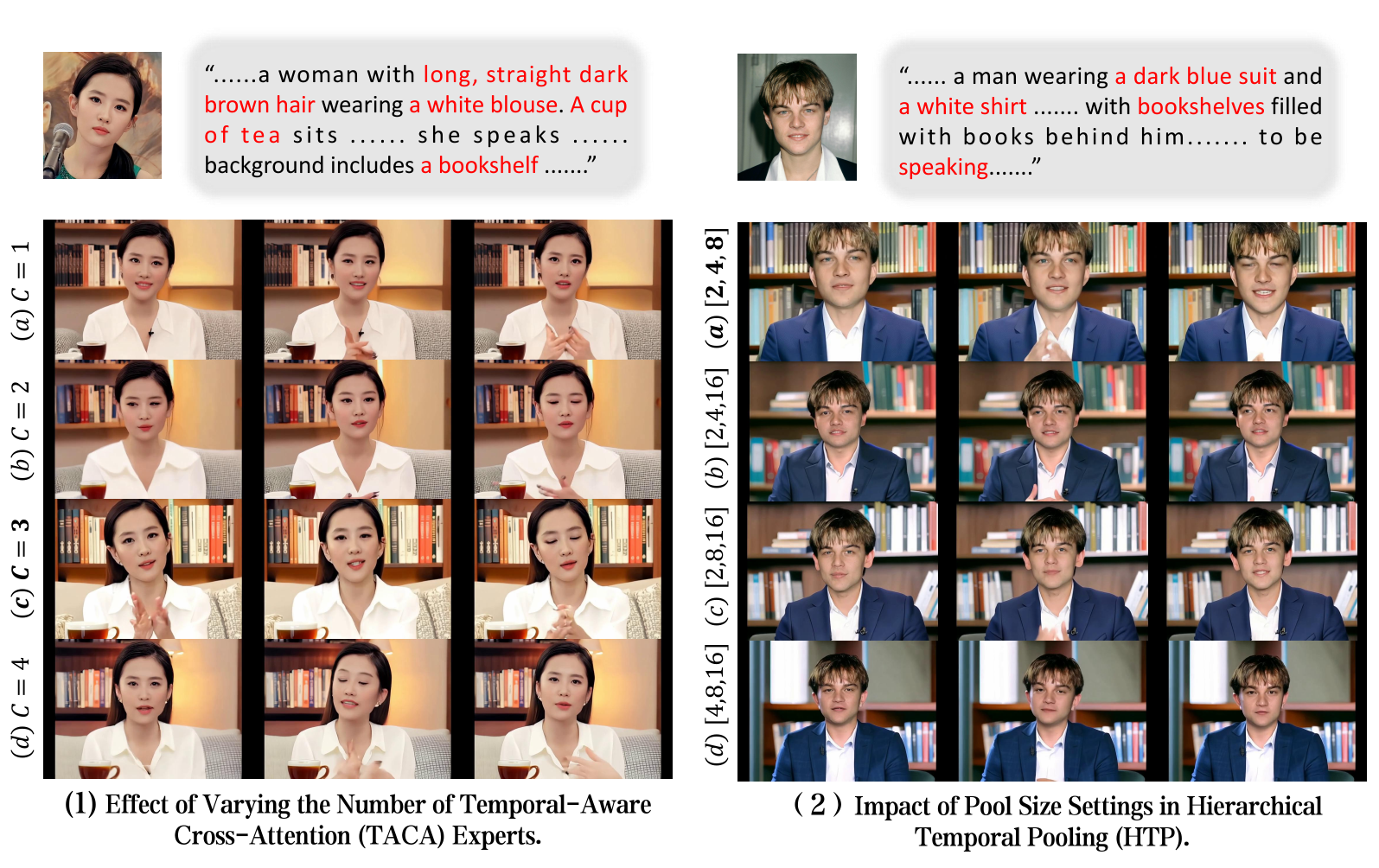}
    \caption{\textbf{Qualitative Results of Hyperparameter Study.} 
    }
    \label{fig:hyperpara}
\end{figure*}

To better demonstrate the effect of different parameter settings in the MoCA layer, we present qualitative results in Figure~\ref{fig:hyperpara}. In subfigure (1), we vary the number of TACA experts in the MoCA layer, and in subfigure (2), we explore the impact of different pool size settings in HTP. As shown in (1), the best facial similarity, video quality, and text alignment are achieved when $C = 3$. In contrast, the results for $C = 1$ and $C = 2$ fail to match the description “long, straight dark brown hair” from the text prompt, and the face similarity noticeably degrades when $C = 4$. Subfigure (2) provides a clear visual comparison of different HTP pool sizes, showing that the best results are obtained when the pool size is set to [2, 4, 8], which is also consistent with the quantitative findings.

\end{document}